\documentclass[11pt, a4paper, twocolumn]{google}

\usepackage[authoryear, sort&compress, round]{natbib}
\usepackage{tikz}
\usetikzlibrary{shapes.geometric, arrows.meta, positioning, fit, backgrounds, calc, decorations.pathreplacing}
\usepackage{xspace}
\usepackage{balance}

\newcommand{\agentgym}{\textsc{Agent~Gym}\xspace}
\newcommand{\alf}{\textsc{ALF}\xspace}


\title{\agentgym: A Framework for Continuous Evaluation and Evolution of LLM Agents Through Human-in-the-Loop Feedback}

\correspondingauthor{pgomran@google.com}

\reportnumber{0001}

\renewcommand{\today}{2026}

\author[1]{Pouya Ghiasnezhad Omran}
\author[1]{Michael Zimmermann}
\author[1]{Duncan Cambridge}
\author[1]{Ashmita Kapoor}
\author[1]{Tanya Dixit}

\affil[1]{Google Cloud}

\uselogo{true}

\begin{abstract}
Large Language Model (LLM) agents deployed in production environments face a fundamental tension: the agent's behavior is frozen at deployment time, while the business rules and edge cases it must handle continue to evolve. Existing approaches address agent construction and one-time evaluation but provide no structured mechanism for continuous post-deployment behavioral correction without modifying the agent's source code. Most of the approaches offered in the market, require intense collection of logs and traces, and re-examining the agent design by the engineering team, a process which is heavy, long and negates the economical value of agentic transformation. We introduce \agentgym, a modular, domain-agnostic framework that wraps any existing LLM-based agent in a continuous evaluation-and-evolution loop. The framework provides six composable capabilities --- Act, Evaluate, Investigate, Correct, Learn, and Observe --- organized across three architectural zones: a constitution layer that codifies domain knowledge in configuration artifacts, a runtime inference pipeline that chains acting, investigation, and adaptive correction, and a learning loop that enables subject matter experts to discover and validate new correction rules through natural language interaction. The key technical contributions include a hybrid deterministic-LLM correction engine with 21 condition operators and three-tier actions, a three-layer investigation architecture for ground-truth-free compliance validation, and a programmatic safety loop that guarantees rule correctness before human approval. We further introduce the Spec-to-Note Gap, an autoencoder-inspired view of agentic system transparency. An open-source reference implementation for invoice processing demonstrates that the framework is fully operational and ready for adoption.
\end{abstract}

\begin{document}

\maketitle

\section{Introduction}
\label{sec:introduction}

LLM-based agents are increasingly deployed to automate complex, rule-governed business processes such as document classification, data extraction, compliance validation, and decision-making~\citep{brown2020language, team2023gemini}. Frameworks such as ReAct~\citep{yao2022react}, LangChain~\citep{chase2022langchain}, and AutoGen~\citep{wu2023autogen} have made it straightforward to build capable agents that achieve impressive initial accuracy. Yet this initial success masks a deeper structural problem that we term the \emph{static agent dilemma}: the agent's behavior is frozen at deployment time, while the business environment it operates in is not.

Three categories of failure emerge in production. First, \emph{edge-case failures}: the agent encounters document configurations or regulatory exceptions absent from its training. Second, \emph{systematic misinterpretations}: the agent consistently misapplies a business rule across an entire category of cases. Third, \emph{drift and rule evolution}: business rules change over time, and the agent has no mechanism to absorb these changes without code modification. We are offering extensible framework, which in the future, can uncover more learning opportunities, like discovering new skills, tools, MCP servers, etc.

The conventional path from ``subject matter expert (SME) identifies a problem'' to ``agent behavior is corrected'' passes through a software engineering bottleneck: the SME describes the issue informally, an engineer interprets the description, modifies agent code or prompts, re-deploys, and the SME verifies. This workflow is slow, lossy, error-prone (i.e. human interactions) and scales poorly~\citep{amershi2019software, sculley2015hidden}.

A further challenge lies in evaluation. Traditional frameworks compare agent output against labeled ground truth~\citep{liang2022holistic, liu2024agentbench}. However, in production, ground truth may not exist for novel cases, and even when reference data is available, it may not capture the nuanced business rules that should govern the decision. What is needed is a dual evaluation model: one that validates against reference data when available, and one that validates decisions against the business rules themselves. The real enteprise world is adverserial to agents, and the agent need to learn from example of one (and not from a well curated dataset).

We introduce \agentgym, a modular framework that addresses these challenges. Our contributions are:

\begin{enumerate}[leftmargin=*]
\item A \textbf{domain-agnostic framework} for continuous agent evaluation and evolution that treats the acting agent as a black box, requiring no modification to the agent's source code. Domain-specific understanding of the agent's pipeline is derived entirely from declarative configuration artifacts (the constitution), not from custom-coded critic, rule-engine,  or learning components; adopting the framework for a new domain requires changing only these artifacts, while all framework code remains unchanged.

\item A \textbf{hybrid deterministic-LLM correction engine} (\alf) with 21 condition operators for deterministic error detection and a three-tier action model for targeted correction.

\item A \textbf{three-layer investigation architecture} for compliance validation without ground truth, employing cached rule discovery and ultra-conservative triple-check verification.

\item A \textbf{programmatic safety loop} for human-in-the-loop rule discovery that guarantees new rules are validated for schema correctness, cross-case impact, and collateral elimination before SME approval. While human-in-the-loop interaction establishes baseline learning, the safety loop actively interrogates proposed rules and solicits clarifying constraints to ensure robust generalization and prevent governance drift.

\item A \textbf{constitution-based governance model} where domain knowledge is captured in declarative configuration artifacts that serve as both human-readable documentation and machine-parseable specifications.

\item The \textbf{Spec-to-Note Gap}, an autoencoder-inspired view of agentic system transparency that generalizes round-trip correctness~\citep{allamanis2024unsupervised} to whole-system granularity.

\item An \textbf{open-source reference implementation\footnote{\url{https://github.com/GoogleCloudPlatform/generative-ai/tree/main/agents/invoice-processing}} } for invoice processing that demonstrates the complete framework is operational and ready for adoption.
\end{enumerate}

The remainder of this paper is organized as follows. Section~\ref{sec:related} positions \agentgym against related work. Section~\ref{sec:framework} presents the framework architecture. Section~\ref{sec:constitution} discusses constitution-based agent creation. Section~\ref{sec:spectonote} introduces the Spec-to-Note Gap. Section~\ref{sec:implementation} describes the reference implementation. Section~\ref{sec:discussion} discusses limitations and future directions. Section~\ref{sec:conclusion} concludes.

\section{Related Work}
\label{sec:related}

\paragraph{LLM Agent Frameworks.}
Recent frameworks have focused on agent \emph{construction}: ReAct~\citep{yao2022react} combines reasoning and acting, AutoGPT~\citep{significant_gravitas2023autogpt} pursues full autonomy, LangChain~\citep{chase2022langchain} provides composable chains, MetaGPT~\citep{hong2024metagpt} assigns roles to agents, and AutoGen~\citep{wu2023autogen} enables multi-agent conversation. These frameworks address how to \emph{build} capable agents but provide no structured mechanism for post-deployment behavioral evolution. \agentgym is complementary: it wraps any agent built with these frameworks in a continuous correction loop.

\paragraph{Agent Evaluation.}
Benchmarks such as HELM~\citep{liang2022holistic}, AgentBench~\citep{liu2024agentbench}, and SWE-bench~\citep{jimenez2024swe} evaluate agent capabilities on static test sets. The LLM-as-judge paradigm~\citep{zheng2023judging} enables scalable evaluation without human labels. These approaches measure agent quality at a point in time but do not address continuous production monitoring or behavioral correction. \agentgym integrates both deterministic and LLM-based evaluation as components of a broader lifecycle.

\paragraph{Self-Improving Agents.}
Self-Refine~\citep{madaan2023self}enables iterative refinement through self-feedback, and Reflexion~\citep{shinn2023reflexion} uses verbal reinforcement learning for agent self-correction. Both approaches modify the agent itself. In contrast, \agentgym leaves the agent untouched, applying corrections in a separate downstream layer. This preserves auditability and enables independent version control of the agent and its correction rules.

\paragraph{Constitutional AI and Alignment.}
Constitutional AI~\citep{bai2022constitutional} uses a set of principles to guide model behavior during training. \agentgym extends the constitutional metaphor to the operational layer: a reconstructed rules book serves as the system's constitution, governing runtime validation and correction rather than model training.

\paragraph{Human-in-the-Loop Learning.}
RLHF~\citep{ouyang2022training, christiano2017deep} and active learning~\citep{settles2009active} incorporate human feedback into model training. \agentgym operates at a different level: Human feedback produces structured correction rules that are applied deterministically at inference time, without retraining the underlying model. This makes corrections immediate, editable, and reversible. In-context and rule-based learning is essential in production settings because knowledge external to model weights remains transparent, auditable, and easily editable. Furthermore, enterprise policy updates and data unlearning can be enacted instantly without incurring the latency and cost of continuous fine-tuning or model retraining.

\paragraph{Business Rule Management.}
Traditional business rule management systems (BRMS) such as Drools provide deterministic rule engines for enterprise applications~\citep{ross2003principles, graham2006business}. \agentgym's \alf engine extends this paradigm with LLM-driven actions, enabling corrections that require understanding of context and semantics beyond what deterministic rules alone can express.

\paragraph{Model Documentation.}
Model Cards~\citep{mitchell2019model} and Datasheets~\citep{gebru2021datasheets} establish standards for documenting ML systems. Microsoft's Transparency Notes~\citep{microsoft2020transparency} extend this to deployed services. The Spec-to-Note Gap concept introduced in Section~\ref{sec:spectonote} proposes automated generation of such documentation, connecting to round-trip correctness~\citep{allamanis2024unsupervised} at the system level.

Table~\ref{tab:comparison} summarizes the positioning of \agentgym relative to key approaches across five dimensions critical for production agent management.

\section{The \agentgym Framework}
\label{sec:framework}

\begin{figure*}[t!]
\centering
\resizebox{\textwidth}{!}{%
\begin{tikzpicture}[
    node distance=0.5cm,
    box/.style={draw, rounded corners=3pt, minimum height=0.9cm, minimum width=1.8cm,
                align=center, font=\scriptsize, line width=0.5pt},
    widebox/.style={draw, rounded corners=3pt, minimum height=1.5cm, align=center,
                    font=\scriptsize, line width=0.5pt},
    stagebox/.style={draw, rounded corners=2pt, minimum height=0.45cm,
                     align=center, font=\tiny, fill=blue!15, inner sep=2pt,
                     line width=0.4pt},
    zonelabel/.style={font=\small\bfseries, anchor=north west, text=brown!70!black},
    arrow/.style={-{Stealth[length=2.5mm]}, thick},
    refarrow/.style={-{Stealth[length=2mm]}, blue!50, dashed, thick},
    feedbackarrow/.style={-{Stealth[length=2.5mm]}, red!60!black, dashed, thick},
    reviewarrow/.style={-{Stealth[length=2.5mm]}, green!50!black, dashed, very thick},
]


\node[box, fill=gray!10] (input) {Case /\\Invoice};

\node[widebox, fill=white, minimum width=7.0cm, minimum height=1.5cm,
      right=0.8cm of input] (actbox) {};
\node[font=\scriptsize\bfseries, anchor=south] at ([yshift=0.02cm]actbox.north) {Acting Agent};

\node[stagebox] at ([xshift=-2.5cm]actbox.center) (s1) {Classify};
\node[stagebox, right=0.1cm of s1] (s2) {Extract};
\node[stagebox, right=0.1cm of s2] (s3) {Validate};
\node[stagebox, right=0.1cm of s3] (s4) {Transform};
\node[stagebox, right=0.1cm of s4] (s5) {Output};
\node[stagebox, right=0.1cm of s5] (s6) {Audit};
\foreach \a/\b in {s1/s2, s2/s3, s3/s4, s4/s5, s5/s6} {
    \draw[-{Stealth[length=1mm]}, thin] (\a) -- (\b);
}

\node[box, fill=blue!22, minimum height=1.5cm, minimum width=2.2cm,
      right=0.8cm of actbox] (invest) {Investigation\\Agent};

\node[font=\tiny, fill=red!18, draw, rounded corners=2pt, inner sep=2pt,
      anchor=center] at ([xshift=-0.45cm]invest.north) (stopbtn) {\textsc{stop}};
\node[font=\tiny, fill=green!12, draw, rounded corners=2pt, inner sep=2pt,
      anchor=center] at ([xshift=0.45cm]invest.north) (contbtn) {\textsc{ok}};

\node[widebox, fill=white, minimum width=2.8cm, minimum height=1.5cm,
      right=0.8cm of invest] (alfbox) {};
\node[font=\scriptsize\bfseries, anchor=south] at ([yshift=0.02cm]alfbox.north) {\alf};
\node[stagebox, fill=gray!12, font=\tiny, inner sep=2pt]
      at ([xshift=-0.55cm]alfbox.center) (rb) {Rule\\[-1pt]Base};
\node[font=\tiny, align=left, anchor=west]
      at ([xshift=0.2cm]alfbox.center) {Collect\\Plan\\Execute};

\node[box, fill=green!15, right=0.8cm of alfbox] (output) {Final\\Output};

\draw[arrow] (input) -- (actbox);
\draw[arrow] (actbox) -- (invest);
\draw[arrow] (invest) -- (alfbox);
\draw[arrow] (alfbox) -- (output);


\node[box, fill=white, draw=gray!50, minimum width=4.0cm, minimum height=1.1cm,
      line width=0.8pt, above=1.8cm of invest]
      (constitution) {\textbf{The Constitution}\\[-2pt]{\tiny Source of Truth \& Transparency Note}};

\draw[refarrow] ([xshift=-0.5cm]constitution.south) -- (invest.north);
\draw[refarrow] ([xshift=0.5cm]constitution.south) -- (alfbox.north);

\draw[refarrow] (actbox.north) -- ([xshift=-1.5cm]constitution.south);


\node[box, fill=orange!10, minimum width=2.4cm,
      below=1.8cm of alfbox] (rla) {Rule Learning\\Agent (RLA)};
\node[box, fill=blue!6, minimum width=2.0cm,
      right=1.2cm of rla] (sme) {Human Expert\\(SME)};

\draw[arrow, dashed] (rla.north) --
      node[right, font=\tiny, pos=0.45, align=left]
      {Update\\[-1pt]Rules} (alfbox.south);

\draw[arrow, dashed] (sme) -- node[above, font=\tiny] {Feedback} (rla);

\draw[arrow, dashed] (output.south) --
      node[right, font=\tiny, pos=0.4, align=left]
      {Sample\\[-1pt]Review} (sme.north);

\draw[reviewarrow] (rla.west) -| ([xshift=-0.4cm]actbox.south west) -- (actbox.south west);
\node[font=\tiny, text=green!40!black, align=center, anchor=east]
      at ([xshift=-0.6cm]rla.west)
      {\textbf{Periodic System Review}\\[-1pt]Promote Frequent Patterns};


\coordinate (L) at ([xshift=-0.6cm]input.west);
\coordinate (R) at ([xshift=0.6cm]sme.east);

\coordinate (z1-top) at ([yshift=0.6cm]constitution.north);
\coordinate (z1z2) at ($(constitution.south)!0.5!(stopbtn.north)$);
\coordinate (z2z3) at ($(output.south)!0.40!(rla.north)$);
\coordinate (z3-bot) at ([yshift=-0.7cm]rla.south);

\begin{scope}[on background layer]
  \fill[blue!5]   (L |- z1-top) rectangle (R |- z1z2);
  \fill[yellow!7] (L |- z1z2)   rectangle (R |- z2z3);
  \fill[green!5]  (L |- z2z3)   rectangle (R |- z3-bot);
\end{scope}

\node[zonelabel] (z1label) at ([xshift=0.15cm, yshift=-0.1cm]L |- z1-top)
     {Zone 1: The Constitution Architecture};
\node[zonelabel] at ([xshift=0.15cm, yshift=-0.1cm]L |- z1z2)
     {Zone 2: The Runtime Inference Pipeline};
\node[zonelabel] at ([xshift=0.15cm, yshift=-0.1cm]L |- z2z3)
     {Zone 3: The Learning \& Evolution Loop};

\node[font=\tiny, anchor=north west, align=left, fill=white, draw=gray!30,
      rounded corners=2pt, inner sep=3pt]
      at ([yshift=-0.08cm]z1label.south west) (legend) {%
\tikz\draw[-{Stealth[length=1.5mm]}, thick] (0,0) -- (0.5,0); Data Flow\\
\tikz\draw[-{Stealth[length=1.5mm]}, blue!50, dashed, thick] (0,0) -- (0.5,0); Reference Line%
};

\end{tikzpicture}%
}
\caption{Three-zone architecture of \agentgym. Zone~1 provides the constitutional foundation (rules book and master data). Zone~2 executes the runtime inference pipeline: Acting Agent $\rightarrow$ Investigation Agent $\rightarrow$ \alf engine. Zone~3 enables human-guided evolution through SME feedback and the Rule Learning Agent. Frequent correction patterns are periodically promoted into the Acting Agent's permanent logic.}
\label{fig:architecture}
\end{figure*}
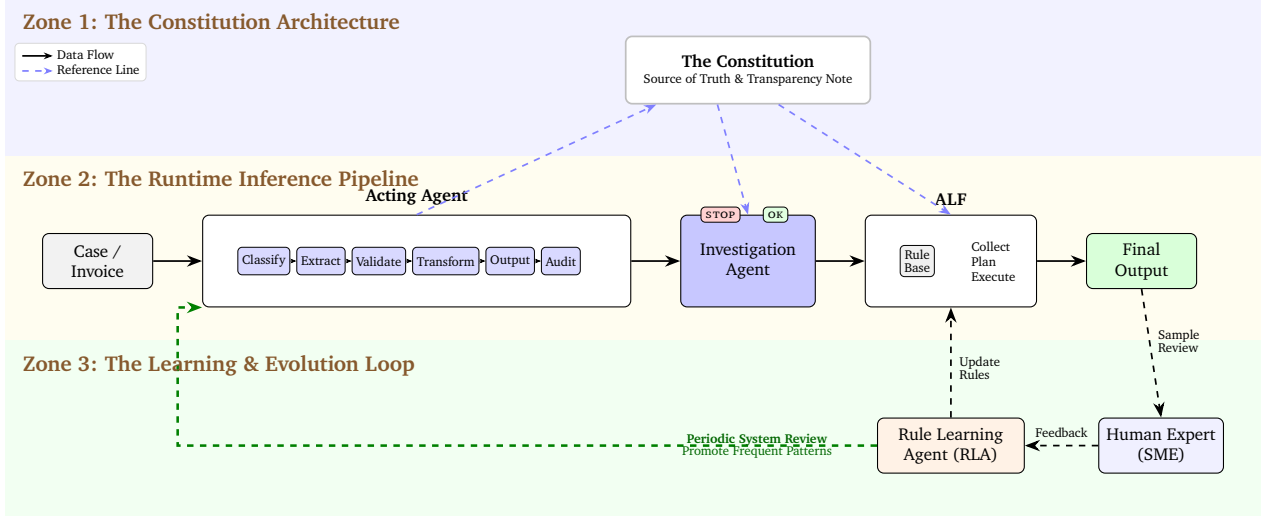

\begin{table*}[t!]
\centering
\caption{Comparison of \agentgym with related approaches across key dimensions for production agent management. \checkmark\ indicates full support; $\sim$ indicates partial support; -- indicates no support.}
\label{tab:comparison}
\small
\begin{tabular}{lccccc}
\toprule
\textbf{Approach} & \textbf{\shortstack{Post-Deploy\\Correction}} & \textbf{\shortstack{Domain\\Agnostic}} & \textbf{\shortstack{Human\\Governance}} & \textbf{\shortstack{No Agent\\Modification}} & \textbf{\shortstack{GT-Free\\Evaluation}} \\
\midrule
ReAct / LangChain / AutoGen & -- & \checkmark & -- & -- & -- \\
HELM / AgentBench / SWE-bench & -- & $\sim$ & -- & \checkmark & -- \\
Self-Refine / Reflexion & $\sim$ & \checkmark & -- & -- & -- \\
RLHF & -- & $\sim$ & \checkmark & -- & -- \\
Traditional BRMS & \checkmark & $\sim$ & $\sim$ & \checkmark & -- \\
\textbf{\agentgym (ours)} & \checkmark & \checkmark & \checkmark & \checkmark & \checkmark \\
\bottomrule
\end{tabular}
\end{table*}

\subsection{Design Principles}
\label{sec:principles}

\agentgym is built on four foundational principles that guide every architectural decision:

\textbf{Principle~1: The agent is a black box.} The framework makes no assumptions about the agent's internal architecture. It observes only input-output behavior: which documents went in, which artifacts came out, and what decisions were rendered. Any agent --- whether a monolithic pipeline, a multi-agent system, or a single LLM call --- can be wrapped without modification.

\textbf{Principle~2: Corrections are layered, not invasive.} Rather than modifying the agent when an error is discovered, corrections are applied in a separate downstream layer. The agent's original output is preserved; the corrected output sits alongside it. This preserves auditability and enables independent rollback.

\textbf{Principle~3: Domain knowledge lives in configuration, not code.} All domain-specific information is captured in a declarative configuration layer. When the framework is adopted for a new domain, only the configuration changes; the framework's code remains untouched.

\textbf{Principle~4: Humans govern the loop.} Every correction rule proposed by the system must be reviewed, approved, and validated by a subject matter expert before it is added to the production rule base (ALF). The framework provides candidate discovery; the human provides authoritative judgment.

\textbf{Principle~5: Governance must be tiered.} To prevent malicious rule injection or unauthorized policy drift, the framework enforces strict separation of duties between runtime rule discovery and constitutional modification. While domain SMEs discover and validate case-level ALF correction rules, permanent modifications to the core constitution and acting agent logic require multi-stakeholder administrative approval.

\subsection{Architecture Overview}
\label{sec:architecture}

\agentgym organizes its capabilities into three architectural zones, illustrated in Figure~\ref{fig:architecture}:

\textbf{Zone~1 --- Constitution Architecture.} A domain configuration layer consisting of a master data specification (YAML) and a reconstructed rules book (Markdown) that together serve as the system's constitution. These artifacts can be authored manually by domain experts or, in the general case, generated by a bootstrap agent --- a four-stage LLM pipeline that analyzes the acting agent's source code and sample outputs. The current reference implementation uses manually authored artifacts; the bootstrap pipeline is a planned automation.  Access to the master data specification (YAML) and reconstructed rules book should be carefully managed to ensure no unauthorized modification as these files wield significant power in critical agents.

\textbf{Zone~2 --- Runtime Inference Pipeline.} Three sequential stages process each case: the acting agent produces initial output, the investigation agent validates compliance against the constitution, and the \alf engine applies targeted corrections based on learned rules.

\textbf{Zone~3 --- Learning and Evolution Loop.} A conversational learning agent enables SMEs to review cases, identify error patterns, and discover new correction rules through a programmatic safety loop that guarantees rule correctness before approval.

The framework provides six composable capabilities that span these zones: Act (the existing agent processes documents), Evaluate (output is compared against ground truth), Investigate (decisions are validated against the rules book), Correct (known error patterns are fixed deterministically), Learn (SMEs discover new rules conversationally), and Observe (operational dashboards, planned for a future release, will provide visibility).

\subsection{Domain Configuration Layer}
\label{sec:config}

The domain configuration layer makes \agentgym domain-agnostic through two generated artifacts:

\textbf{Master Data} is a structured YAML document that fully describes a domain across eleven sections: document types, extraction schemas, taxonomies, validation pipeline definitions, output schema, evaluation comparison groups, investigation file maps, configuration defaults, domain-specific detection logic, artifact naming conventions, and rejection templates. Every framework component reads from this single source of truth.

\textbf{Rules Book} is a human-readable Markdown document that captures the complete business rules governing the agent's processing logic. It describes every validation step, threshold, rejection template, and decision outcome. The investigation agent consumes this document as context for LLM-based validation. We think this is a major innovation area, where the agent DNA is exposed (and editable) not only for engineers, but for domain experts (which are far from code editing and complex configuration files)

In the general case, both artifacts can be generated by a \emph{Bootstrap Agent}, a four-stage LLM pipeline that: (1)~analyzes the acting agent's source code to extract pipeline structure, schemas, and validation logic; (2)~scans sample outputs to discover artifact schemas and status values; (3)~synthesizes a comprehensive rules book; and (4)~produces the structured master data YAML. Alternatively, domain experts can author the artifacts directly. The current reference implementation uses manually authored artifacts; the bootstrap pipeline is a planned automation that will reduce onboarding effort for new domains.

Adopting \agentgym for a new domain requires providing the acting agent's code and sample outputs, along with manually authored or bootstrap-generated configuration artifacts.

\subsection{Runtime Inference Pipeline}
\label{sec:pipeline}

\subsubsection{Acting Agent}
\label{sec:acting}

The acting agent is the system under observation. \agentgym imposes only one structural requirement: the agent must produce \textbf{JSON artifacts} as output. Beyond this, the agent's architecture is unconstrained. The acting agent processes input documents and produces a folder of output artifacts per case, including intermediate results and a final structured output file containing the agent's decision.

\subsubsection{Investigation Agent}
\label{sec:investigation}

The Investigation Agent validates agent decisions against the rules book \emph{without requiring any ground truth data}, making it applicable to every case, including novel ones.

The agent employs a three-layer validation architecture, summarized in Table~\ref{tab:investigation}:

\textbf{Layer~1 --- Deterministic Checks.} At each case, the agent performs data source validation, verifying that the acting agent drew field values from the correct sources (extraction vs.\ preprocessing). Bypass detection identifies cases where the agent skipped required validation steps. These checks involve no LLM calls.

\textbf{Layer~2 --- LLM Rule Discovery.} At initialization, the rules book is sent to an LLM, which discovers and categorizes validation rules into structured groups. The discovered rules are cached with a SHA-256 hash of the rules book content, so subsequent runs skip rediscovery unless the rules book has changed.

\textbf{Layer~3 --- Conservative Cross-Validation.} An ultra-conservative layer evaluates each rule group against the case. When a potential violation is detected, a \emph{triple-check mechanism} re-runs the evaluation twice more. Only violations confirmed by all three independent runs are reported. This dramatically reduces false positives.

\paragraph{Overhead control.}
Four mechanisms bound the investigation agent's LLM costs in practice. \emph{Content-hash caching} ensures that Layer~2 computes a SHA-256 hash of the rules book content; on subsequent runs the cached rule groups are reused unless the rules book has changed, amortizing Layer~2's LLM cost to zero after the first invocation. \emph{Section-filtered context} reduces prompt size by extracting only the rules book section relevant to the current validation phase rather than sending the full document to the LLM for each call. \emph{Batch grouping} consolidates deterministic rule groups (data-source priorities, tolerance thresholds, entity whitelists, and keyword lists) into a single batched LLM call; only subjective groups such as work-type classification receive individual calls. Finally, \emph{early exit from triple-check} issues the second and third verification calls only when the first call detects a potential violation. For compliant cases, which constitute the majority under a well-functioning agent, Layer~3 incurs a single LLM call per rule group. Together, these mechanisms ensure that steady-state investigation cost is dominated by Layer~1 deterministic checks, with LLM usage concentrated on the initial run and on genuine violations.

The investigation agent produces per-case compliance scores on a 0--100\% scale, categorized as \textsc{Fully\_Compliant} ($\geq$80\%), \textsc{Partial\_Violation} (60--80\%), or \textsc{Major\_Violation} ($<$60\%). A major violation halts the pipeline, preventing potentially non-compliant output from reaching the correction stage.

\begin{table}[t]
\centering
\caption{Three-layer investigation architecture. Each layer adds progressively deeper validation with controlled LLM usage.}
\label{tab:investigation}
\small
\begin{tabular}{llcc}
\toprule
\textbf{Layer} & \textbf{Method} & \textbf{LLM} & \textbf{Scope} \\
\midrule
1 & Deterministic & No & Per-field \\
2 & Rule discovery & Yes$^*$ & Global \\
3 & Cross-validation & Yes & Per-group \\
\bottomrule
\multicolumn{4}{l}{\footnotesize $^*$Cached by content hash; amortized to zero after first run.}
\end{tabular}
\end{table}

\subsubsection{ALF --- Adaptive Learning Framework}
\label{sec:alf}

\alf is the correction engine that sits downstream of the acting agent. Its design separates \emph{detection} from \emph{correction}: detection is fully deterministic, while correction leverages LLMs when contextual understanding is required.

\paragraph{Detection.}
Each \alf rule specifies conditions evaluated against the agent's output using a library of 21~operators, including equality, containment, regex matching, numeric comparison, list membership, null checks, prefix matching, and dynamic field references. Conditions are joined with AND logic. There is no LLM involvement in detection --- a rule either matches or it does not, and the result is perfectly reproducible.

Dynamic field references (e.g., {\small\texttt{\_DYNAMIC\_\allowbreak{}preprocessing.\allowbreak{}vendor\_name\_}}) are resolved at evaluation time, enabling parameterized conditions that adapt to case-specific context.

\paragraph{Correction.}
When a rule matches, its action specifies how to correct the output. \alf supports a three-tier action model, summarized in Table~\ref{tab:alf_tiers}:

\begin{table}[t]
\centering
\caption{\alf action tiers. Tier selection depends on the nature and extent of the error.}
\label{tab:alf_tiers}
\small
\begin{tabular}{clcl}
\toprule
\textbf{Tier} & \textbf{Action} & \textbf{LLM} & \textbf{Scope} \\
\midrule
1 & Field edit & No & Single field \\
2 & Surgical patch & Yes & Target fields \\
3 & Pipeline cont. & Yes & Full output \\
\bottomrule
\end{tabular}
\end{table}

Tier~1 applies deterministic field edits for cases where the correct value is a known constant. Tier~2 invokes an LLM to determine correct values for specific fields, patching them into the existing output. Tier~3 is used when the agent terminated early (e.g., rejecting at an initial phase when it should have continued); the LLM continues the pipeline from the resumption point, producing a complete revised output.

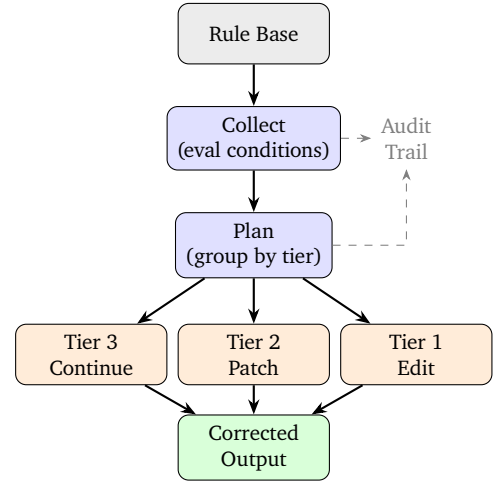
\begin{figure}[t]
\centering
\begin{tikzpicture}[
    node distance=0.55cm,
    box/.style={draw, rounded corners, minimum height=0.8cm, minimum width=2.0cm, align=center, font=\scriptsize},
    arrow/.style={-{Stealth[length=2mm]}, thick},
]
\node[box, fill=gray!15] (rulebase) {Rule Base};
\node[box, fill=blue!12, below=of rulebase] (collect) {Collect\\(eval conditions)};
\node[box, fill=blue!12, below=of collect] (plan) {Plan\\(group by tier)};

\node[box, fill=orange!15, below left=0.6cm and 0.1cm of plan] (t3) {Tier 3\\Continue};
\node[box, fill=orange!15, below=0.6cm of plan] (t2) {Tier 2\\Patch};
\node[box, fill=orange!15, below right=0.6cm and 0.1cm of plan] (t1) {Tier 1\\Edit};
\node[box, fill=green!15, below=1.8cm of plan] (out) {Corrected\\Output};

\draw[arrow] (rulebase) -- (collect);
\draw[arrow] (collect) -- (plan);
\draw[arrow] (plan) -- (t3);
\draw[arrow] (plan) -- (t2);
\draw[arrow] (plan) -- (t1);
\draw[arrow] (t3) -- (out);
\draw[arrow] (t2) -- (out);
\draw[arrow] (t1) -- (out);

\node[right=0.4cm of collect, font=\scriptsize, text=gray, align=center] (audit) {Audit\\Trail};
\draw[-{Stealth[length=1.5mm]}, gray, dashed] (collect.east) -- (audit);
\draw[-{Stealth[length=1.5mm]}, gray, dashed] (plan.east) -| (audit);

\end{tikzpicture}
\caption{\alf Collect-Plan-Execute pipeline. Conditions are evaluated deterministically (Collect), actions are grouped by tier (Plan), and executed in order: Tier~3 $\rightarrow$ Tier~2 $\rightarrow$ Tier~1. A full audit trail is maintained throughout.}
\label{fig:alf}
\end{figure}

\paragraph{Aggregation.}
When multiple rules match, \alf uses a \emph{Collect-Plan-Execute} pipeline, illustrated in Figure~\ref{fig:alf}. All matching rules are collected, merged into a revision plan that combines actions by tier, and executed in fixed order: pipeline continuation first, surgical patches second, deterministic edits last. Scope-based mutual exclusion guarantees at most one rule fires per scope, preventing conflicts. This ensures at most two LLM calls per case regardless of how many rules match.

Each correction is fully audited: which rules were evaluated, which matched, the revision plan, and complete LLM metadata (model, token counts, latency).

\subsection{Learning and Evolution Loop}
\label{sec:learning}

The Learning Agent enables SMEs to review agent output, describe desired corrections in natural language, and collaboratively discover new \alf rules. It orchestrates four sub-modules:

\textbf{Case Loader} presents the acting agent's output for a specific case, showing the decision, phase-by-phase validation results, extracted data, and correction history.

\textbf{Rule Discoverer} accepts the SME's natural language feedback and interprets it in the context of the case data, existing rules, and the rules book. Using an LLM grounded in these artifacts, it generates candidate conditions following a \emph{conservative domain principle}: conditions should be narrow enough to match only the intended cases but general enough to cover similar future cases.

\textbf{Impact Assessor} evaluates the proposed rule deterministically against a sample of existing cases (with the target case always included), reporting target matches (the case under review must match), collateral matches (other cases that also match), and safe non-matches.

\textbf{Rule Writer} validates the schema, checks for conflicts with existing rules, backs up the rule base, and persists the approved rule with full metadata: who approved it, when, which cases it was designed for, and which rules book section it relates to. Each learning session is logged for auditability.

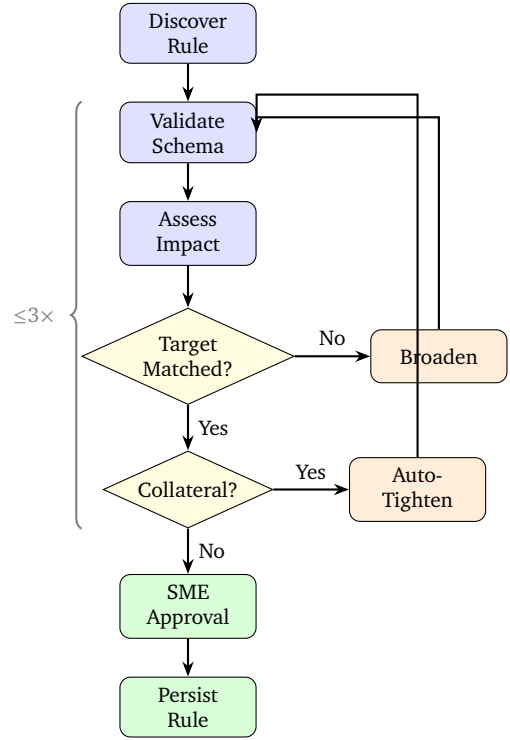
\begin{figure}[t]
\centering
\begin{tikzpicture}[
    node distance=0.5cm,
    box/.style={draw, rounded corners, minimum height=0.7cm, minimum width=1.8cm,
                align=center, font=\scriptsize},
    decision/.style={draw, diamond, aspect=2.2, minimum height=0.7cm,
                     align=center, font=\scriptsize, inner sep=2pt},
    arrow/.style={-{Stealth[length=2mm]}, thick},
]
\node[box, fill=blue!12] (discover) {Discover\\Rule};
\node[box, fill=blue!12, below=of discover] (validate) {Validate\\Schema};
\node[box, fill=blue!12, below=of validate] (impact) {Assess\\Impact};
\node[decision, fill=yellow!15, below=0.55cm of impact] (targetcheck) {Target\\Matched?};
\node[box, fill=orange!15, right=1.0cm of targetcheck] (broaden) {Broaden};
\node[decision, fill=yellow!15, below=0.6cm of targetcheck] (check) {Collateral?};
\node[box, fill=orange!15, right=1.0cm of check] (tighten) {Auto-\\Tighten};
\node[box, fill=green!15, below=0.6cm of check] (approve) {SME\\Approval};
\node[box, fill=green!15, below=of approve] (persist) {Persist\\Rule};

\draw[arrow] (discover) -- (validate);
\draw[arrow] (validate) -- (impact);
\draw[arrow] (impact) -- (targetcheck);
\draw[arrow] (targetcheck) -- node[right, font=\scriptsize] {Yes} (check);
\draw[arrow] (targetcheck) -- node[above, font=\scriptsize] {No} (broaden);
\draw[arrow] (broaden.north) |- ([yshift=0.2cm]validate.east) -- (validate.east);
\draw[arrow] (check) -- node[right, font=\scriptsize] {No} (approve);
\draw[arrow] (check) -- node[above, font=\scriptsize] {Yes} (tighten);
\draw[arrow] (tighten.north) |- ([yshift=0.5cm]validate.east) -- (validate.east);
\draw[arrow] (approve) -- (persist);

\draw[gray, thick, decorate, decoration={brace, amplitude=4pt, mirror}]
    ([xshift=-0.5cm]validate.north west) --
    ([xshift=-0.5cm]validate.north west |- check.south)
    node[midway, left=6pt, font=\scriptsize, text=gray] {$\leq$3$\times$};

\end{tikzpicture}
\caption{Programmatic safety loop for rule discovery. Candidate rules are iteratively validated, checked for target coverage, and auto-tightened to eliminate collateral matches before SME approval.}
\label{fig:safety}
\end{figure}

The critical innovation is the \emph{programmatic safety loop}, illustrated in Figure~\ref{fig:safety}. This loop is enforced in code --- not prompt instructions --- so it cannot be bypassed by the LLM. Each candidate rule passes through an iterative validation cycle:

\begin{enumerate}[leftmargin=*]
\item \textbf{Schema validation}: The rule's JSON structure is verified against the \alf schema.
\item \textbf{Target match verification}: The rule's conditions are evaluated against the target case. If the rule fails to match, the LLM automatically broadens conditions while remaining conservative.
\item \textbf{Collateral assessment}: Conditions are evaluated deterministically against a sample of cases. If unintended matches are found, the LLM automatically adds narrowing conditions (vendor name, amount range, service category) to eliminate them.
\end{enumerate}

This cycle repeats up to three times. If collateral persists after three attempts, the rule is presented to the SME with a warning. The SME may also request iterative revisions to a proposed rule; each revision re-enters the safety loop. At no point is a rule applied without explicit human approval. Care must be taken to control access to the approval process for rule persistence and in some cases a two step approval process may be required with an administrator or senior manager providing approval, after SME approval, before a rule is persisted.

\paragraph{Rule lifecycle management.}
As the rule base grows, systematic lifecycle management becomes necessary to prevent it from becoming unwieldy. The framework provides several structural supports for this. Each rule carries an \texttt{enabled} flag that allows deactivation without deletion, preserving audit history while removing the rule from runtime evaluation. The Rule Writer creates a timestamped backup of the entire rule base before any modification, enabling rollback of any change. Before a new rule is persisted, conflict detection checks for duplicate identifiers, priority collisions within the same scope, and scope overlap with existing rules. Structured metadata fields (severity, root cause, rules-book section) enable rules to be queried, filtered, and prioritized for review. Rules can also be permanently deleted, with a backup created before removal. Beyond these building blocks, the architecture anticipates a periodic system review cycle, illustrated in Figure~\ref{fig:architecture}. When a correction rule fires so consistently that it reveals a systematic deficiency in the acting agent, the underlying pattern should be promoted into the agent's permanent logic, and the corresponding \alf rule retired. This promotion path keeps the rule base lean by ensuring that well-established corrections graduate out of the runtime correction layer and into the agent itself. The current implementation does not yet automate this promotion, nor does it include review queues, performance tracking per rule, or sunset policies; concrete mechanisms for these capabilities are discussed in Section~\ref{sec:discussion}.

\subsection{Evaluation Engine}
\label{sec:evaluation}

The Evaluation Engine provides quantitative accuracy measurement when ground truth is available. It operates as a schema-driven comparison framework with two layers:

\textbf{Deterministic comparison} performs field-by-field matching across configured comparison groups, applying financial tolerances (default: \$0.02) for numeric fields and exact matching for text fields after normalization.

\textbf{LLM-as-judge}~\citep{zheng2023judging} (optional) provides a holistic alignment verdict per case: \textsc{Aligned}, \textsc{Partially\_Aligned}, or \textsc{Not\_Aligned}. This captures semantic alignment beyond what field-level comparison can express.

Both layers read their configuration from the master data, ensuring that adding a new comparison group requires only a YAML change, not a code change.

\section{Constitution-Based Agent Creation}
\label{sec:constitution}

A distinctive property of \agentgym's architecture is that the constitutional artifacts --- the rules book and master data --- can serve not only as governance instruments for an existing agent but also as \emph{specifications for creating new agents}. We outline this bidirectional relationship between constitution and agent.

\paragraph{Constitution as specification.}
The rules book captures every validation step, threshold, rejection condition, and decision outcome in human-readable form. The master data YAML codifies document types, extraction schemas, taxonomies, and pipeline configurations. Together, these artifacts constitute a complete, machine-parseable specification of the desired agent behavior. A developer building a new acting agent can use these artifacts as requirements documents, and an LLM can use them as grounding context for code generation.

\paragraph{Bootstrap and refinement cycle.}
A bootstrap agent can establish an initial constitution from an existing agent's code. Alternatively --- and as demonstrated in the reference implementation --- the constitution can be authored first by a domain expert describing the desired processing logic, then used to guide agent construction. This inverts the conventional flow: rather than code $\rightarrow$ constitution, the flow becomes constitution $\rightarrow$ code $\rightarrow$ refined constitution.

\paragraph{Co-evolution.}
As the learning loop produces correction rules that address systematic agent errors, patterns emerge that indicate where the acting agent's logic should be updated. The constitution captures these patterns in structured form. Over time, the constitution evolves to reflect not just the original design intent but also the accumulated operational experience encoded in correction rules. This positions the constitution as a living document that bridges the gap between initial specification and deployed behavior, providing a foundation for principled agent re-engineering when code-level changes become warranted.

\paragraph{Onboarding new agents.}
Adopting \agentgym for a new acting agent requires two inputs: the acting agent itself and a constitution that describes its intended behavior. The framework's computational components, including the \alf engine, the investigation agent, and the learning agent, are domain-independent and require no modification; they operate entirely from the constitutional artifacts. A \emph{bootstrap agent}, itself an LLM-based pipeline, can automate constitution generation by analyzing the new agent's source code and sample outputs. Because agent frameworks such as ADK impose consistent structural conventions, the bootstrap agent can extract pipeline stages, extraction schemas, validation logic, and decision rules from the code, then synthesize a reconstructed rules book and master data YAML without manual authoring. This bootstrap pipeline is described in Section~\ref{sec:config} and is planned as a future automation; the current reference implementation uses manually authored artifacts. Once the constitution is in place, the \alf rule base starts empty. Subject matter experts then populate it through the learning loop as they observe the new agent's behavior, discover error patterns, and validate correction rules through the programmatic safety loop described in Section~\ref{sec:learning}. This separation ensures that the framework scales to new domains and new agents without engineering effort beyond constitution generation.

This bidirectional relationship distinguishes \agentgym from approaches that treat agent development and agent monitoring as separate concerns. The framework's configuration artifacts are simultaneously governance instruments and development specifications.

\section{The Spec-to-Note Gap: An Autoencoder View of Agentic System Transparency}
\label{sec:spectonote}

Modern agentic systems are rarely a single model. They are compositions of planners, tools, retrievers, guardrails, and sometimes multiple agents managing control flow~\citep{guo2024large, dorri2018multiagent}. Documenting what such a system \emph{actually does} by hand is tedious and goes stale immediately. We propose a structural pattern that addresses this challenge.

\subsection{The Autoencoder Analogy}

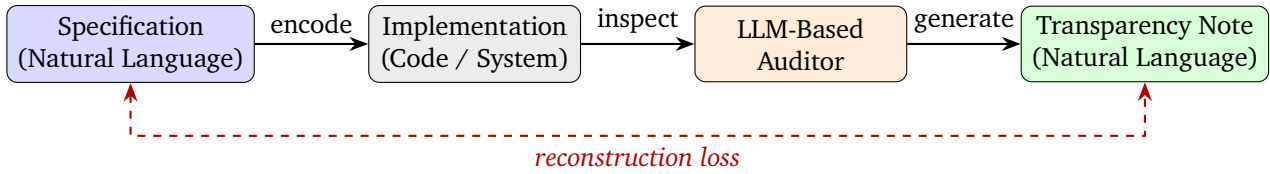
\begin{figure*}[t]
\centering
\begin{tikzpicture}[
    node distance=0.8cm,
    box/.style={draw, rounded corners, minimum height=1.0cm, minimum width=2.8cm, align=center, font=\small},
    arrow/.style={-{Stealth[length=2.5mm]}, thick},
]
\node[box, fill=blue!15] (spec) {Specification\\(Natural Language)};
\node[box, fill=gray!15, right=1.5cm of spec] (code) {Implementation\\(Code / System)};
\node[box, fill=orange!15, right=1.5cm of code] (auditor) {LLM-Based\\Auditor};
\node[box, fill=green!15, right=1.5cm of auditor] (note) {Transparency Note\\(Natural Language)};

\draw[arrow] (spec) -- node[above, font=\small] {encode} (code);
\draw[arrow] (code) -- node[above, font=\small] {inspect} (auditor);
\draw[arrow] (auditor) -- node[above, font=\small] {generate} (note);

\draw[{Stealth[length=2.5mm]}-{Stealth[length=2.5mm]}, thick, red!70!black, dashed]
    (spec.south) -- ++(0,-0.7) -| node[below, pos=0.25, font=\small, text=red!70!black] {\textit{reconstruction loss}} (note.south);

\end{tikzpicture}
\caption{The Spec-to-Note autoencoder. Natural language specification is encoded into an implemented system, then decoded back to natural language by an LLM auditor. The gap between spec and transparency note surfaces missing or unintended behaviors.}
\label{fig:autoencoder}
\end{figure*}

Consider the lifecycle of an agentic system as a transformation chain, illustrated in Figure~\ref{fig:autoencoder}:

\begin{enumerate}[leftmargin=*]
\item A \textbf{specification} (natural language) describes the intended behavior.
\item \textbf{Implementation} (code, prompts, tools, configuration) encodes this spec into an executable system.
\item An \textbf{LLM-based auditor} inspects the implementation artifacts --- code, prompt templates, tool definitions, evaluation results, sample execution traces --- and generates a \textbf{transparency note} in natural language.
\end{enumerate}

This chain forms an autoencoder over natural language: the spec is the input, the implemented system is the latent representation, and the transparency note is the reconstruction. The analogy is not merely metaphorical --- it yields a concrete diagnostic: \emph{contrasting the spec against the transparency note functions as a reconstruction loss}, surfacing missing capabilities, silent scope creep, and behaviors the evaluation suite never measured.

\subsection{Connections to Prior Work}

This pattern generalizes the round-trip correctness idea of \citet{allamanis2024unsupervised} from function-level consistency to whole-system granularity. It also connects to verbal feedback loops such as Self-Refine~\citep{madaan2023self} and Reflexion~\citep{shinn2023reflexion}, with two key differences: the artifact being critiqued is a \emph{system} rather than a single output, and the critique is itself a \emph{deliverable} --- a transparency note in the tradition of Model Cards~\citep{mitchell2019model} and Datasheets~\citep{gebru2021datasheets}.

\subsection{SME Interface}

The transparency note serves as a shared interface with subject matter experts. SMEs cannot read evaluation harnesses, and even when they can, evaluations only check what the developers thought to measure. But they can read a structured natural-language note about how the system behaves in their domain, and they will quickly flag wrong assumptions, missed populations, and regulatory edge cases. Comments on the note become tickets on the system.

Within \agentgym, the Spec-to-Note pattern maps naturally to the Observe capability. An auditor agent can be run on a CI cadence over the system's artifacts, regenerating the transparency note as the constitution and rule base evolve.

\subsection{Open Questions}

Several questions warrant further investigation: how to measure the auditor's own faithfulness to the system it inspects; when to regenerate the note (on every commit, on every rule change, on a fixed cadence); whether the gap between spec and note can be turned into an actual optimization signal rather than a review aid; and whether an adversarially-prompted auditor could serve as a release-gate red-teamer.

\section{Reference Implementation}
\label{sec:implementation}

We have developed an open-source reference implementation that demonstrates the complete \agentgym framework in the domain of invoice processing. The implementation is built on Google's Agent Development Kit (ADK) and uses Gemini~\citep{team2023gemini} models for all LLM operations.

\subsection{System Overview}

The reference implementation is a unified dual-mode agent --- a single \texttt{LlmAgent} instance with 18~registered function tools that supports both inference (document processing) and learning (SME-guided rule discovery) modes. The entire system is packaged as a self-contained Python module: all data, configurations, test cases, rules, and evaluation artifacts reside within the package, enabling deployment without external infrastructure dependencies.

\subsection{Acting Pipeline}

The acting agent implements a nine-stage sequential pipeline:
\begin{enumerate}[leftmargin=*]
\item \textbf{Classifier}: Identifies document types (invoice, work authorization form, email).
\item \textbf{Extractor}: Pulls structured data from PDFs using LLM-based extraction with Pydantic schema validation~\citep{wang2023grammar}.
\item \textbf{Phase~1--4 Validators}: Progressive compliance validation --- intake checks, content validation, external validation (e.g., tax ID checksum), and calculation validation.
\item \textbf{Transformer}: Normalizes line items against standard taxonomies.
\item \textbf{Output Generator}: Produces the final structured decision.
\item \textbf{Audit Logger}: Creates a compliance trail.
\end{enumerate}

Each stage produces a numbered JSON artifact (e.g., \texttt{01\_classification.json} through \texttt{09\_audit\_log.json}), providing full traceability. An early-exit mechanism allows the pipeline to skip remaining validation phases when a rejection is determined, proceeding directly to output generation.

\subsection{Investigation and Correction}

The investigation agent implements the three-layer architecture described in Section~\ref{sec:investigation}. Layer~2 caches discovered rules using SHA-256 hashing of the rules book content, eliminating redundant LLM calls across runs. Layer~3's triple-check mechanism uses confidence thresholds of 90\% for violations and 70\% for ambiguous cases, treating ambiguity as compliant (conservative bias). Deterministic rule groups are batched into a single LLM call, and only the rules-book section relevant to each validation phase is included in the prompt rather than the full document, further reducing per-case token consumption.

The \alf engine implements all 21~condition operators described in Section~\ref{sec:alf}, including dynamic field references that resolve values like {\small\texttt{\_DYNAMIC\_\allowbreak{}preprocessing.\allowbreak{}vendor\_name\_}} at evaluation time. The Collect-Plan-Execute pipeline handles multi-rule scenarios with scope-based mutual exclusion.

\subsection{Learning Mode}

The learning mode implements the full safety loop (Section~\ref{sec:learning}). The rule discoverer generates candidate rules grounded in the rules book and existing rule base. The impact assessor evaluates conditions deterministically against a sample of cases (always including the target). Auto-tightening adds conditions based on vendor name, amount range, rejection template text, and service category to eliminate collateral matches.

\subsection{Evidence of Framework Viability}

The reference implementation includes a comprehensive test suite covering condition operators, action executors, schema validation, conflict detection, and impact assessment is publicly available\footnote{\url{https://github.com/GoogleCloudPlatform/generative-ai/tree/main/agents/invoice-processing}} for invoice processing use case. A two-layer evaluation framework (deterministic field comparison with configurable tolerances, plus optional LLM-as-judge) enables systematic quality measurement.

The implementation demonstrates several properties that validate the framework design:
\begin{itemize}[leftmargin=*]
\item \textbf{Domain adaptability}: Replacing the master data YAML and acting pipeline adapts the entire framework to a new document type. All downstream components --- evaluation, investigation, \alf, and learning --- automatically adjust.
\item \textbf{Operational readiness}: The system runs via ADK's web interface or command line, with documented paths to production deployment on cloud infrastructure.
\item \textbf{Self-containment}: The complete system, including test cases with PDF documents, ground truth, and pre-configured rules, ships as a single installable package.
\end{itemize}

\section{Discussion and Future Work}
\label{sec:discussion}

\paragraph{Strengths.}
\agentgym's principal strength is its separation of concerns: the acting agent, the correction layer, and the learning loop are independently versioned, tested, and evolved. The domain-agnostic design, achieved through the configuration layer, means the framework code is reusable across business processes. The human-in-the-loop governance model ensures that corrections are traceable and reversible.

\paragraph{Limitations.}
The current framework has several limitations that future work should address. First, the bootstrap process is currently a manual workflow; an LLM-assisted bootstrap agent, as described in Section~\ref{sec:config}, would reduce the effort required to author constitutional artifacts for new domains. Second, the investigation agent's LLM-based validation incurs costs that scale with the number of cases and rule groups, though the caching, batching, section filtering, and early-exit mechanisms described in Section~\ref{sec:investigation} substantially reduce this in practice; quantifying the cost reduction across large-scale deployments remains future work. Third, our reference implementation validates the framework in a single domain; multi-domain validation is needed to fully establish domain agnosticism.

\paragraph{Future directions.}
Several extensions follow naturally from the current architecture:

\emph{Automated rule suggestion.} As the investigation agent accumulates compliance data, recurring violation patterns can be identified and proactively surfaced to the SME, shifting the learning agent from reactive to proactive operation.

\emph{Multi-run trend analysis.} An administrative dashboard could support loading multiple agent runs, enabling comparative analysis across configurations and time periods.

\emph{Rule lifecycle management.} As the rule base accumulates corrections over time, several capabilities become necessary. \emph{Periodic review}, referenced in Figure~\ref{fig:architecture} as ``Periodic System Review'', would flag rules that have not matched any case within a configurable time window for SME review and potential deprecation. \emph{Performance tracking} would record per-rule match counts and correction outcomes, and whether the corrected output was subsequently validated as correct by the evaluation engine, surfacing rules with high match rates but poor outcomes for revision. \emph{Consolidation} would identify clusters of rules with overlapping conditions that could be merged into a single, more general rule, reducing rule-base size without sacrificing coverage. Most importantly, \emph{promotion into the acting agent} would address cases where a correction rule fires so consistently that it reveals a systematic agent deficiency; the corresponding logic should be promoted into the acting agent's permanent behavior and the \alf rule retired. Currently, the learning agent modifies only the \alf rule base; closing this loop is an important direction for future work.

\emph{Cross-domain transfer.} Organizations operating multiple agents across different business processes could share a single \agentgym installation, with domain-specific behavior governed entirely by configuration. Effective rule structures and condition patterns discovered in one domain may transfer to others.

\emph{Automated bootstrap.} The bootstrap agent described in Sections~\ref{sec:config} and~\ref{sec:constitution} would enable fully automated onboarding of new acting agents by analyzing their source code and sample outputs to generate the constitutional artifacts. Investigating how to make this pipeline robust across different agent frameworks and how to handle agents whose internal structure does not follow conventional framework patterns are important open questions.

\emph{Spec-to-Note as release gate.} The autoencoder pattern described in Section~\ref{sec:spectonote} could be integrated into CI/CD pipelines, with the reconstruction loss serving as an automated quality gate for agent deployments.

\section{Conclusion}
\label{sec:conclusion}

We have presented \agentgym, a modular framework for continuous evaluation and evolution of LLM agents through human-in-the-loop feedback. The framework addresses the static agent dilemma --- the fundamental tension between frozen agent behavior and evolving business environments --- by wrapping any existing agent in a structured observation-correction-learning loop without modifying the agent's source code.

The framework's technical contributions include a hybrid deterministic-LLM correction engine that separates reliable detection from flexible correction, a three-layer investigation architecture that validates compliance without ground truth, and a programmatic safety loop that guarantees rule correctness before human approval. The constitution-based governance model captures domain knowledge in declarative artifacts that serve simultaneously as human documentation and machine specifications. The Spec-to-Note Gap concept provides a principled approach to automated system transparency.

An open-source reference implementation demonstrates that the framework is not merely theoretical but fully operational. \agentgym does not replace the need for well-engineered agents; it provides the environment in which those agents can be continuously measured, understood, and improved --- not by engineers rewriting code, but by domain experts contributing the knowledge that only they possess.

\section*{Acknowledgments}
We would like to thank Mitesh Agarwal for his valuable feedback and for reviewing this paper.

\section*{Declaration on Generative AI}
During the preparation of this work, the author(s) used LLM-based
tools in order to: assist with drafting, perform grammar and spell checks, assist with LaTeX formatting and bibliography
management. After using these tools, the author(s) reviewed and
edited the content as needed and take full responsibility for the
publication's content.
\balance
\bibliography{main}

\end{document}